\documentclass{article} 
\usepackage[preprint]{colm2026_conference}

\usepackage{latexsym}

\usepackage[T1]{fontenc}
\usepackage{microtype}
\usepackage{hyperref}
\usepackage{url}
\usepackage{booktabs}
\usepackage{microtype}
\usepackage[most]{tcolorbox}
\usepackage{inconsolata}

\usepackage{graphicx}

\usepackage{minitoc}
\usepackage{xcolor}
\usepackage{algorithm}
\usepackage{algpseudocode}
\usepackage{amsmath}
\usepackage{setspace}
\usepackage{mdframed}
\usepackage{amssymb} 
\usepackage{tcolorbox}
\tcbuselibrary{listings,breakable}
\usepackage{listings}
\usepackage{inconsolata}
\usepackage{adjustbox}
\usepackage{booktabs}
\usepackage{makecell}
\usepackage{multicol}
\usepackage{multirow}

\usepackage{lineno}

\definecolor{math}{HTML}{4B8BBE}      
\definecolor{coding}{HTML}{E69F00}    
\definecolor{knowledge}{HTML}{8F7E4F} 
\definecolor{tooluse}{HTML}{D55E00}   
\definecolor{feedbackcolor}{HTML}{1f3c73} 
\definecolor{selffeedcolor}{HTML}{4a90e2} 
\definecolor{bexefeedbackcolor}{HTML}{a6c8ff} 
\definecolor{selmv}{HTML}{e74c3c} 
\definecolor{icl}{HTML}{e8d9c4} 
\definecolor{dc}{HTML}{c6a682} 
\definecolor{nofed}{HTML}{7a5a3f} 

\newcommand{\eg}{\textit{e.g.}}

\newcommand{\ie}{\textit{i.e.}}

\title{Chain-of-Experience for Continual LLM Improvement}

\author{Haoqin Tu$^{1,2}$*\quad Yunhao Fang$^{2}$*\quad Yizhong Wang$^{2}$\quad Cihang Xie$^{1}$\quad Shen Yan$^{2}$\\
\small
$^1$ \textbf{UC Santa Cruz}\quad $^2$ \textbf{Bytedance Seed}\quad \\ 
\small
* equal contribution and work done at Bytedance Seed
}

\begin{document}

\maketitle
\begin{abstract}
Humans continuously learn from experience, whereas conventional large language model (LLM) evaluations ignore the models' ability to improve through inference-time interaction.
In this paper, we study how LLMs learn from iterative experience at test time, a setting we refer to as Chain-of-Experience (CoE), where models accumulate experiential traces through iterative interactions with self or environmental feedback to form a continual improvement loop beyond zero-shot inference.
We instantiate CoE with diverse feedback mechanisms, including model self-feedback and environmental signals such as correctness or public coding test pass rates, and evaluate across math, coding, and knowledge domains using 8 LLMs, including GPT-5, Gemini-2.5 Pro, Claude-4.5 Sonnet.
Our study shows that leveraging iterative experience consistently outperforms feedback-free baselines, achieving substantial gains with self feedback alone, alongside a 5.6\% overall improvement and 19\% lower API cost across tasks and models. We further show that combining complementary feedback channels (\eg, model and correctness signals) yields additional gains, and that CoE delivers higher accuracy per token than existing test-time strategies.
We observe a positive correlation between LLM base ability and improvement capacity, and show that models remain robust under weak or spurious feedback, with different feedback contributing to distinct improvement aspects and most gains emerging early in the iterations.
\end{abstract}

\section{Introduction}
Humans naturally and continuously learn from their experiences, each success or failure contributes to an evolving understanding that informs future decisions.  
Contemporary machine learning systems --- modern large language models (LLMs) in particular~\citep{achiam2023gpt,team2023gemini,grattafiori2024llama,liu2024deepseek,yang2025qwen3} --- behave quite differently: once trained, they are deployed in a fixed state, treating every inference as an isolated event and ignoring the rich feedback embedded in the problem solving process itself. 
In contrast, learning should be a continuous process, LLMs should be able to improve not only during training but also at test-time by iteratively interacting with their environment --- processing feedback, engaging in reflection to constantly update their understanding, or shortly, improving from their own knowledge and experience~\citep{silver2025welcome,snell2025scaling}. 

Inspired by successes in reinforcement learning, search-based methods explore a vast range of solutions.
Heuristic strategies~\citep{yao2023tree,hao2023reasoning}, including majority voting~\citep{wang2022self,wang2024boosting} aim to identify higher-quality answers leveraging verifiers from candidates parallelly generated by LMs.
Nevertheless, the model experience from these methods is still temporary, which is usually consolidated into a \textit{single answer} and then discarded, leaving models to restart each problem without accumulated insight.
An alternative paradigm is self-refinement, where models are enabled to iteratively improve through self-critique and correction~\citep{madaan2023self,shinn2023reflexion}, internalizing feedback within a single context.
Extensions such as self-debugging~\citep{chen2023teaching} and Reflexion~\cite{shinn2023reflexion} incorporate external signals like code execution signals, but they remain fragmented and shallow in the use of iterative experience.

Building on these attempts, we conceptualize Chain-of-Experience (CoE), wherein models engage in iterative problem-solving with feedback, and investigate: \textit{``How can LLMs evolve from accumulated experience with interactions and feedback to improve during test time?''}
We use the CoE framework to systematically explore prolonged environment interactions across four feedback types: none, model feedback, executor feedback for code tasks, and correctness feedback for general tasks.

\begin{figure*}[t]
  \centering
  \includegraphics[width=\textwidth]{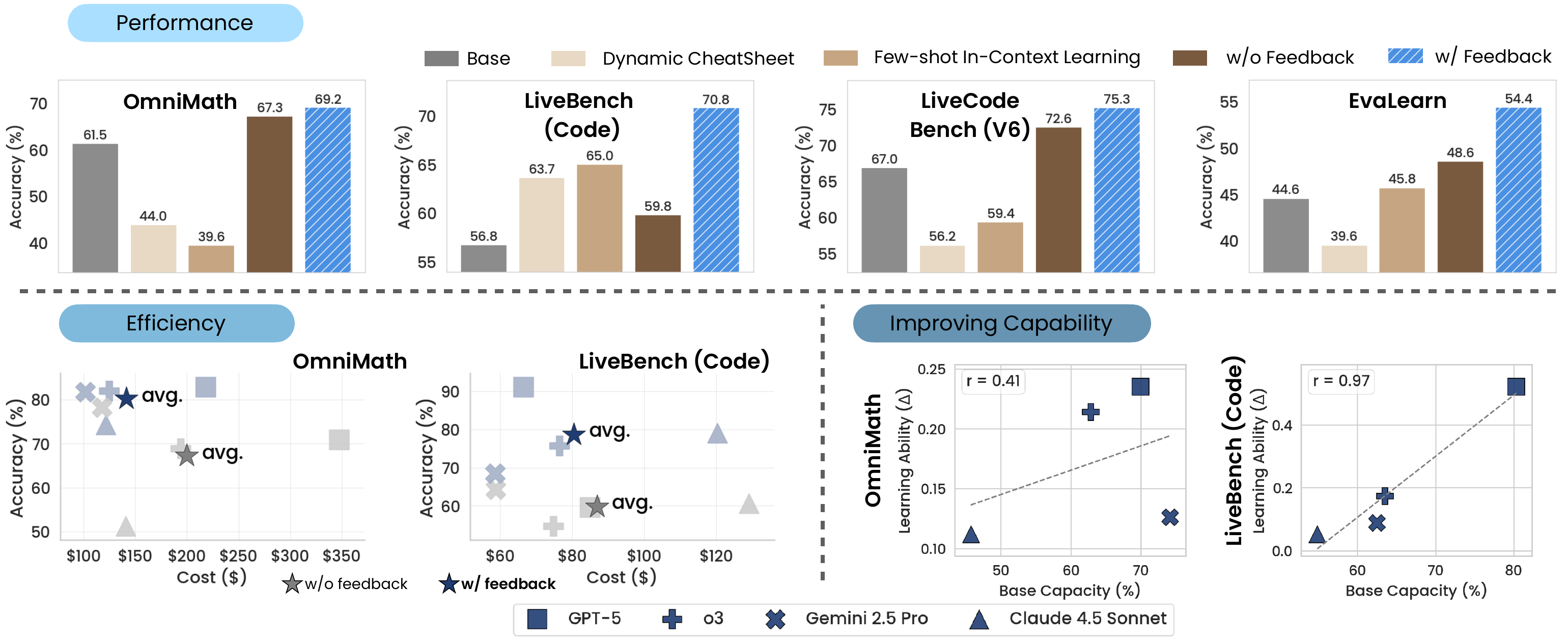}
  \vspace{-5mm}
\caption{Summarized results on four benchmarks across math, code, and knowledge over four LLMs. Iterative problem solving under CoE provides three benefits: \emph{upper:} by incorporating \textcolor{feedbackcolor}{\textbf{feedback}} in CoE, the average performance across four LLMs outperforms other test-time augmentations; \emph{lower left:} models are capable of achieving the better performance with lower API cost with \textcolor{feedbackcolor}{\textbf{feedback}}; \emph{lower right:} LLMs that perform better on the task display better improving capabilities through CoE with moderate to strong Pearson correlation. We present more explanations regarding baselines in Section~\ref{sec:baselines}.}
\label{fig:teaser_results}
\end{figure*}

To evaluate CoE-based algorithms, we prompt eight state-of-the-art models, including GPT-5, o3, Gemini-2.5 Pro, Claude-4.5 Sonnet, across math, coding, and knowledge domains. Extensive experiments over four feedback types—drawn from existing algorithms and our designed paradigm (\ie, none, executor, model, and correctness feedback)—show that CoE with feedback consistently yields notable improvements (Figure~\ref{fig:teaser_results}).
In particular, simple CoE methods substantially outperform test-time scaling approaches that rely on experience from other tasks~\citep{suzgun2025dynamic,zhang2025agentic}, achieving an average of 7-9\% gain over these algorithms with just self feedback (average 62.9\% to 71.0\%). Moreover, models leveraging iterative experience and feedback reason more efficiently, delivering a 5.6\% overall improvement with 19\% lower API cost and higher accuracy per token across all tasks and models. Combining complementary feedback channels (\eg, model and correctness/executor signals) yields further improvements, while memory-based selection methods applied within task do not outperform full experience trails, suggesting that aggressive compression may discard critical intermediate reasoning.
We further observe a clear positive correlation between learning gains from feedback and base ability, with an average Pearson correlation of +0.5 across five benchmarks, indicating that stronger models evolve more effectively from experience.
Finally, our analyses reveal deeper behavioral insights: models remain robust under spurious or weak feedback, most gains occur early in the iterations, and distinct improvement trajectories emerge under different feedback types.

\section{Related Work}
\label{app:related_work}
\paragraph{Training-free Test-time Strategies.}
Large language models can reason without additional training via various test-time prompting strategies.
Chain-of-Thought (CoT) elicits step-by-step reasoning and improves arithmetic, commonsense, and symbolic tasks~\citep{wei2022chain}, spawning a family of ``Chain-of-X'' methods~\citep{yu2023chain,li2023chain,huang2023chain,chia2023contrastive}.
Representative variants include contrastive CoT~\citep{chia2023contrastive}, least-to-most prompting~\citep{zhou2022least}, task-specialized forms such as Chain-of-Explanation, Chain-of-Note, and Chain-of-Knowledge~\citep{huang2023chain,yu2023chain,li2023chain}, and Tree-of-Thought (ToT), which explores multiple reasoning paths via search and outperforms CoT on planning tasks~\citep{yao2023tree}.
With the emergence of large reasoning models such as OpenAI’s \textit{o} series~\citep{jaech2024openai} and DeepSeek R1~\citep{guo2025deepseek}, verifier-based methods that select among parallel generations have regained attention, including step-level~\citep{lightman2023let,wang2024math,zhang2024rest} and output-level~\citep{zheng2023judging,cai2024internlm2,liu2025skywork} verification.
While effective as post-processing using external feedback~\citep{liu2024skywork,wang2025visualprm,tu2025vilbench,wang2024helpsteer}, these methods lack the iterative loop for model evolving.
Our CoE is also training-free, but differs by using feedback to drive iterative self-evolving during inference.

\paragraph{Learning from Experiences.}
Learning from experience underlies both human intelligence and AI systems.
Reinforcement learning formalizes experience through policy gradients and actor--critic methods~\citep{williams1992simple,schulman2017proximal}, achieving success in games, robotics, and control~\citep{silver2016mastering}, and more recently via post-training on online generations to improve alignment and reasoning~\citep{bai2022training,guan2024deliberative,shao2024deepseekmath,yu2025dapo}.
Beyond training, experience can accumulate during inference.
For cross-task experience, Dynamic CheatSheet (DC)~\citep{suzgun2025dynamic}, Agentic Context Engineering (ACE)~\citep{zhang2025agentic}, and related approaches~\citep{zheng2023synapse,wang2024agent,zhao2024expel} maintain persistent inference-time memories that distill reusable strategies, while agentic scaffolds enable collective experience sharing across agents~\citep{tang2025agent,chen2025swe,ouyang2025reasoningbank,han2026vlaa}.
For same-task experience, Reflexion~\citep{shinn2023reflexion}, Self-Refine~\citep{madaan2023self}, Self-Debug~\citep{chen2023teaching}, S*~\citep{li2025s}, Iteration-of-Thought~\citep{radha2024iteration}, and ReVeal~\citep{jin2025reveal} iteratively refine outputs using self-feedback or execution signals, while recent pipelines further exploit offline model experience for agent improvement~\citep{zhang2025agent,chen2025scaling}. Most recent works~\citep{xia2026metaclaw,tu2026visualclaw,} also extend such paradigm to real-world usage.
In contrast to prior work, we present a unified framework that treats a model’s entire solving history as experience and systematically studies diverse feedback signals to enable continual improvement at test time.

\section{Model Improvement via CoE}
In this section, we first provide general concepts of the iterative problem solving setup, termed \textit{Chain-of-Experience} (CoE), followed by a detailed discussion of four diverse feedback types to enhance model experience. 
Finally, we provide explanations on how we scale up the iteration of experience to probe model learning performance at test-time.

\subsection{Overview}
In the traditional question-answering setting~\citep{sutskever2014sequence,raffel2020exploring,roberts2020much,brown2020language}, when given a question $Q$, large language models (LLMs) will generate a plausible response $A$ sampled from the conditional distribution $P(A \mid Q)$. 
To extend this paradigm into the era of experience~\citep{silver2025welcome}, we incorporate an environment feedback variable $F$ to represent the observable consequence or evaluation of responses when grounded in an interactive environment. 
Formally, the feedback $f$ is sampled from the conditional distribution $f\sim P'(F \mid Q,A)$, where $P'$ is modeled by an environment that can be instantiated as a coding execution environment, an internal world model of the agent, or even a real-world environment, which provides generative~\citep{ouyang2022training,mahan2024generative} or environment feedback~\citep{shao2024deepseekmath,madaan2023self} for the given sequence.

In this paper, we investigate a paradigm that extends the single-turn formulation into a sequential decision process, where each response $a_i$ at step $i$ depends on the full history and corresponding environmental feedback of prior attempts ; we refer to this setting as \textit{Chain-of-Experience} (CoE).
The generative process is defined:
\begin{equation}
    a_t \sim P(a_t\mid Q,e_0,e_1,\ldots,e_{t-1})
\end{equation}
where $e_i$ is the $i^{th}$ experience consists of $(a_i, f_i)$.

\begin{figure*}[t]
  \centering
  \includegraphics[width=0.8\textwidth]{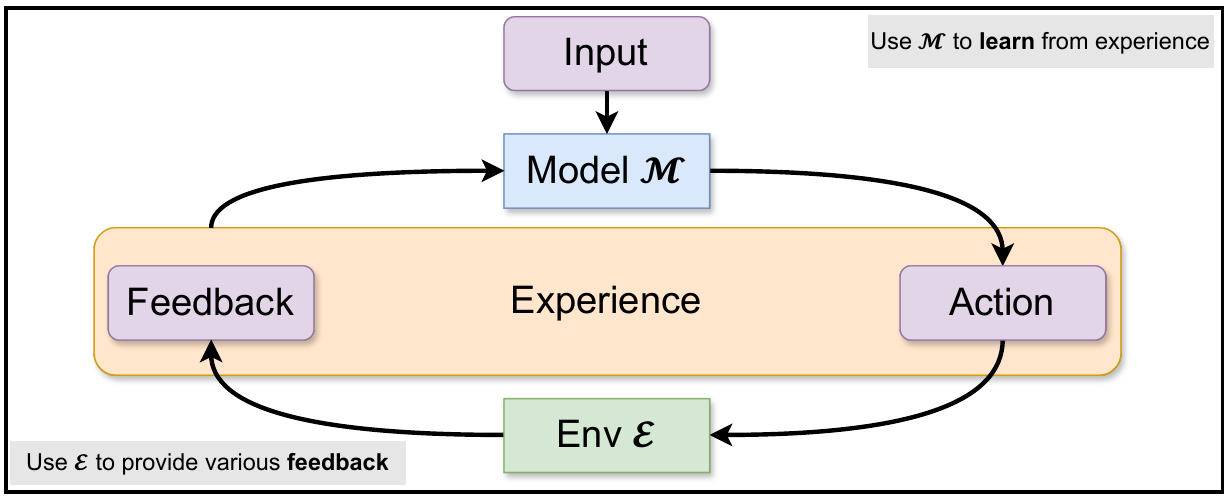}
\caption{An overview of our studies iterative improvement loop for LMs. The model $\mathcal{M}$ learns by repeatedly interacting with the environment $\mathcal{E}$ (\eg, model simulation or coding environment). In each cycle, $\mathcal{M}$ generates actions conditioned on the input and accumulated experience, then receives feedback from $\mathcal{E}$ to form a new experience. 
We investigate four specific feedback in this study: none, execution, model, and correctness feedback.}
\end{figure*}

\subsection{Feedback Spectrum}

To fully characterize how different forms of experience shape the model’s iterative evolution, we categorize feedback along a spectrum of richness --- from completely implicit to strongly explicit signals. 
Each feedback type corresponds to a specific instantiation of the environment's response, which in turn influences the next-step action via the CoE generative process.
In general, the model can be updated according to 
\[
    a_t \sim P(a_t \mid Q, (a_0,f_0), \ldots, (a_{t-1},f_{t-1})),
\]
where the function $f_i$ is the feedback we employed at the $i-$th iteration.
Below, we describe these four feedback types used in this study, together with their formal definitions.

\noindent \textbf{No feedback.}
The environment provides no evaluation or signal, \ie, $f_i = \varnothing$.
The experience reduces to $e_i = (a_i, \varnothing)$, the next action is sampled from
\[
    a_t \sim P(a_t \mid Q, a_0, a_1, \ldots, a_{t-1}),
\]
meaning any improvement must arise from reflection on prior attempts, without external guidance.

\noindent \textbf{Execution feedback.}
For tasks grounded in executable or interactive environments (\eg, coding tasks with interpreters or unit tests), feedback is generated by running the model’s response $a_i$ inside the environment $\mathcal{E}$:
\[
    f_i \sim P'(F \mid Q, a_i) = \mathcal{E}(Q, a_i),
\]
where $f_i$ may include execution traces, error messages, runtime logs, or test-case outcomes. 

\noindent \textbf{Model feedback.}
A (possibly identical) auxiliary language model $\mathcal{M}_{\text{fb}}$ acts as a judge or critic $f_i = \mathcal{M}_{\text{fb}}(Q, a_i)$,
where $f_i$ may include textual critiques, preference scores, or structured evaluations. 
This enables refinement even in the absence of an external environment, relying purely on linguistic or preference-based signals.

\noindent \textbf{Correctness feedback.}
When a domain-specific verifier is available, the environment supplies binary correctness signals:
\[
    f_i = \mathbf{1}\{a_i \text{ is correct}\} \in \{0,1\}.
\]
Such oracle-like information provides explicit fine-grained evaluation of success and failure.
Although this type of feedback is often unrealistic in real-world settings, where ground-truth verification is costly or unavailable, still, we include it as a high-signal reference setting to approximate an upper bound on the benefits of iterative refinement.


\vspace{-.5em}
\section{Experiments}
\subsection{Experiment Setup} \label{sec:baselines}
\noindent \textbf{Datasets.}
We focus on three different tasks: \textcolor{math}{math}, \textcolor{coding}{coding}, and \textcolor{knowledge}{knowledge}.
More specific, we select two benchmarks for each task: \textcolor{math}{AIME 2025}~\citep{balunovic_srimatharena_2025}, \textcolor{math}{OmniMath}~\citep{gao2024omni}, \textcolor{coding}{LiveCodebench (V6)}~\citep{jain2024livecodebench}, \textcolor{coding}{LiveBench (Code)}~\citep{white2024livebench}, \textcolor{knowledge}{EvaLearn}~\citep{dou2025evalearn}, and \textcolor{knowledge}{GPQA Diamond}~\citep{rein2024gpqa}. Detailed descriptions are in Appendix~\ref{app:eval_bmks}.

\noindent \textbf{Baselines.}
As for baselines, we examine model skills in either (1) utilizing different levels of built-in reasoning or (2) leveraging experience from previously solved problems. 
For controlling reasoning depth, OpenAI and Claude models can be tuned to produce varying amounts of reasoning tokens. For methods that absorb experience from prior examples, we adopt few-shot in-context learning (ICL)~\citep{brown2020language} as a standard baseline. For a more sophisticated approach, we select Dynamic CheatSheet~\citep{suzgun2025dynamic} and Agentic Context Engineering (ACE)~\citep{zhang2025agentic}, which maintain a continually updated external memory of reusable strategies. Although follow-up works~\citep{ouyang2025reasoningbank,cai2025flex} introduce finer-grained refinements in a similar processing loop, we use these two as the representative baseline. We select the most $k\in [1, 5, 8, 12, 15, 20]$ relevant solutions for ICL, DC, and ACE to form their context, we present baseline details in Appendix~\ref{app:baselines}.

\noindent \textbf{Models.}
We focus on the latest language models with inherent reasoning abilities from various developers to probe their improving capabilities during test-time: GPT-5~\citep{openai2025_gpt5}, GPT-5-mini~\citep{openai2025_gpt5}, o4-mini~\citep{openai2025_o3}, o3~\citep{openai2025_o3}, o3-mini~\citep{openai2025_o3}, Gemini-2.5 Pro~\cite{comanici2025gemini}, Claude 4.5 Sonnet~\cite{anthropic_2025_45}. We run all experiments for three times and report the mean and standard deviation statistics. We present detailed prompting configurations in Appendix~\ref{app:exp_setup}.

\subsection{Scaling with Experience} \label{sec:exp_scaling}
\vspace{-3mm}
\begin{figure*}[t]
  \centering
  \includegraphics[width=\textwidth]{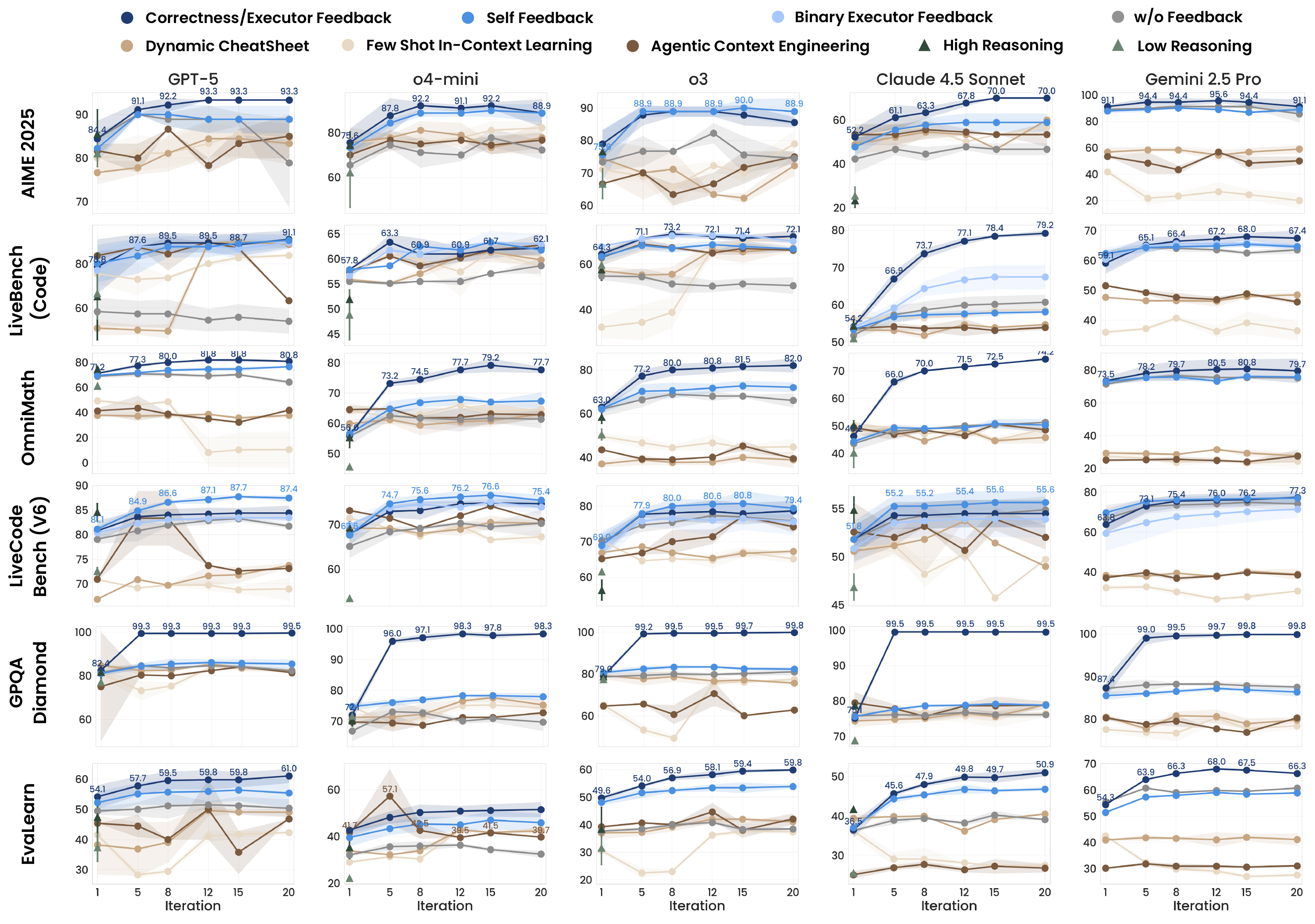}
\caption{Results of five state-of-the-art LLMs on six benchmarks using different generation techniques. Models under CoE with different levels of feedback (\textcolor{feedbackcolor}{correctness/executor}, \textcolor{selffeedcolor}{self}, and \textcolor{bexefeedbackcolor}{binary executor} feedback) generally perform better than the baseline strategies (\textcolor{nofed}{no feedback}, \textcolor{dc}{DC}, \textcolor{icl}{ICL}, and the ones with different \textcolor{gray}{reasoning efforts}). Results are averaged over 3 runs and we shade the standard deviation with a lighter color and plot bars showing model performance under different reasoning efforts. The full results are in Appendix~\ref{app:full_price}.}
\label{fig:main}
\end{figure*}

\hspace{0.1mm}
\begin{mdframed}[backgroundcolor=gray!15] 
\noindent\textbf{Findings 1}: Performance: The Chain-of-Experience setting with feedback boosts reasoning LLMs on various tasks.
\end{mdframed}

LLMs equipped with feedback consistently outperform almost all baselines and settings. As shown in Figure~\ref{fig:main}, across six benchmarks, the Chain-of-Experience (CoE) paradigm—where models iteratively absorb and reuse prior feedback—delivers substantial performance gains (full results in Appendix~\ref{app:full_price}). With self feedback or executor/correctness feedback (as an upper bound), eight modern reasoning models achieve average improvements of 5.6\% and 11.1\% over their no-feedback counterparts, underscoring the value of explicit outcome-based signals for model refinement.
For coding-centric tasks such as LiveBench (Code) and LiveCodeBench (V6), programmatic executor feedback derived from public test-case verification drives sharp accuracy gains of 8.6\% on average (from 66.4\% to 75.0\%), while self-judgement feedback still provides a 7.0\% lift. This indicates that models can internalize both fine-grained and abstract feedback cues into subsequent reasoning. On non-coding tasks (\eg, AIME 2025, OmniMath, GPQA Diamond), the same trend holds: correctness feedback establishes an upper bound, and self feedback—though noisier—continues to foster improvement (\eg, 75.1\% $>$ 67.1\% $>$ 62.5\% w/o feedback).

In comparison, although ICL, DC, and ACE remain widely used, none demonstrates reliable scaling under advanced reasoning models. Averaged across six benchmarks in Table~\ref{tab:feedback_comparison_full}, ICL, ACE, and DC achieve only 62.1\%, 64.0\%, and 62.7\% respectively, all trailing a simple without baseline (66.8\%). In contrast, incorporating explicit feedback yields consistent gains: results with model self feedback reaches 71.0\% (+7-9\% over ICL/ACE/DC), while the best feedback signal further improves performance to 79.3\%. 
Overall, these results proves that feedback-driven CoE acts as a more general and effective test-time scaling framework, enabling LLMs to autonomously improve across domains over no-feedback inference and other baselines.

\begin{figure*}[t]
  \centering
  \includegraphics[width=.9\textwidth]{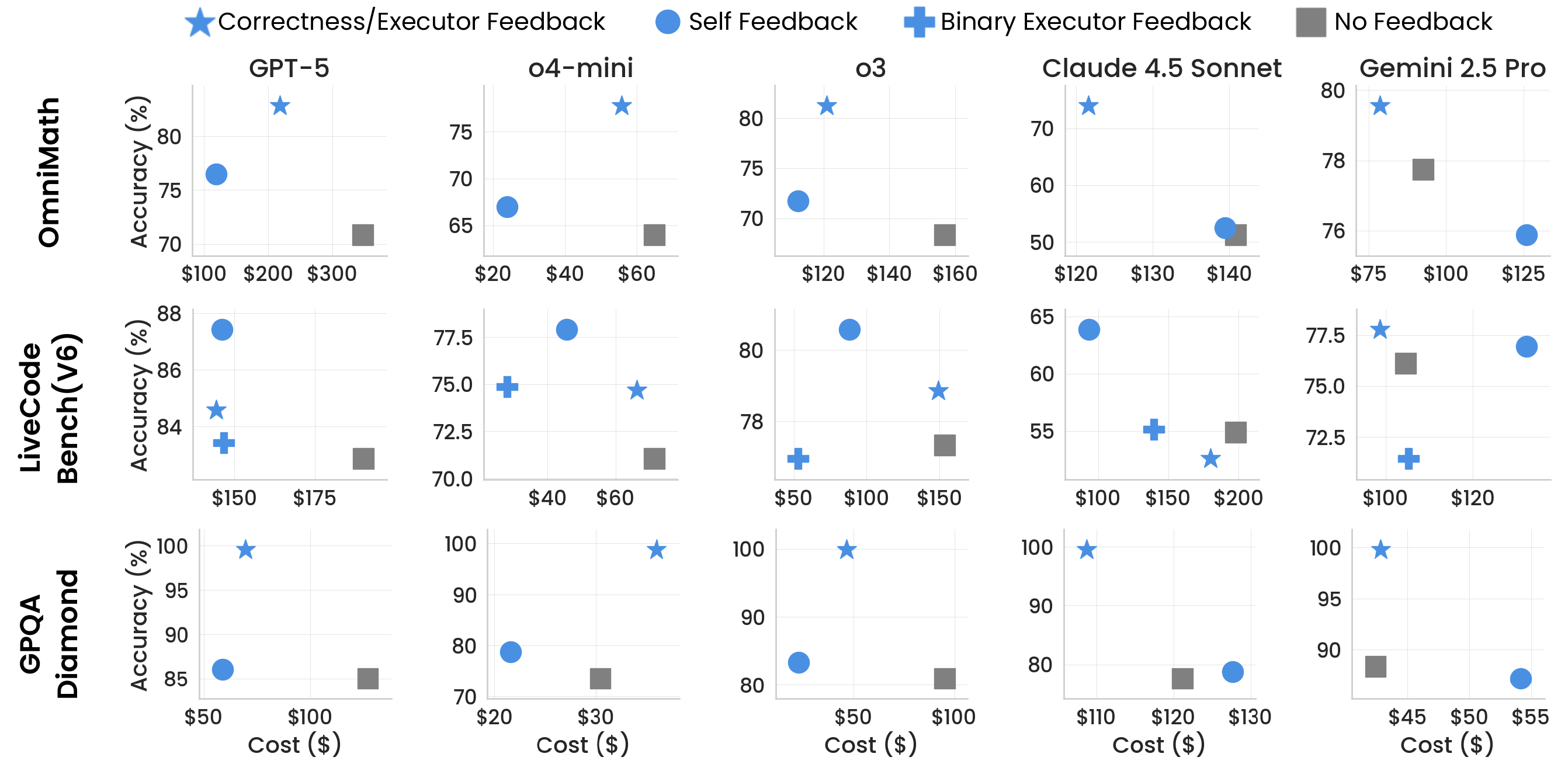}
\caption{Total cost of each model over task completion vs. its best performance within 20 iterations. LLMs with \textcolor{selffeedcolor}{feedback} generally achieve higher scores with fewer costs (at the upper left), while iterative experience without feedback generally falls behind (at lower right). We provide full results over six benchmarks in Appendix~\ref{app:full_price}. }
\label{fig:price}
\end{figure*}

\begin{mdframed}[backgroundcolor=gray!15] 
\noindent\textbf{Findings 2}: Efficiency: LLMs with feedback strikes a balance between performance and API calling cost.
\end{mdframed}

As shown in Figure~\ref{fig:price}, incorporating feedback into CoE not only enhances performance but also improves efficiency --- models with feedback tend to achieve higher accuracy at lower API costs (upper left of sub-figures). Across tasks, most LLMs using feedback-based variants (blue objects) consistently dominate the no feedback baselines (gray square), except Gemini 2.5 Pro. Full results are in Figure~\ref{fig:full_price}, Appendix~\ref{app:full_price}.
For the two coding tasks, \textcolor{selffeedcolor}{self feedback} emerges as a cost-effective compromise: it captures the majority of the performance gain of \textcolor{feedbackcolor}{executor} (\ie, 73.4\% vs. 75.0\%) while incurring 13.4\% fewer API calls (\eg, \$70.7 vs. \$81.6), and moreover, requires 20\% less cost than the \textcolor{gray}{no-feedback} counterpart. 
Similarly, on AIME 2025 and EvaLearn, \textcolor{selffeedcolor}{self feedback} provides more informative input with substantially lower API cost with 47.3\% and 7.0\% reductions across all eight LLMs, even compared with the \textcolor{gray}{no-feedback} solution (\eg, \$8.8 vs. \$4.6 on AIME~25 and \$325.3 vs. \$302.4 on EvaLearn), while still yielding an average 4.4\% and 6.9\% accuracy improvement, respectively.
One exception is Gemini 2.5 Pro, whose self-feedback produces more verbose output with smaller gains, suggesting it may be less effective at self-evaluation than other LLMs.
Beyond API cost, we further analyze token-level efficiency in Table~\ref{tab:token_performance} (Appendix~\ref{app:token_analysis}). CoE with correctness/executor feedback achieves the best accuracy with total token counts comparable to other multi-round methods, yielding substantially higher return per token. For example, on AIME 2025, correctness feedback reaches 84.6\% with 108K tokens, while DC uses only 11K tokens yet achieves a lower 74.7\%, and no-feedback CoE consumes similar tokens (107K) but trails at 74.1\%. This pattern holds consistently across benchmarks, confirming that CoE reallocates compute into feedback iterations and generates less verbose outputs, rather than simply inflating prompt length.


\begin{figure*}[t]
  \centering
  \includegraphics[width=\textwidth]{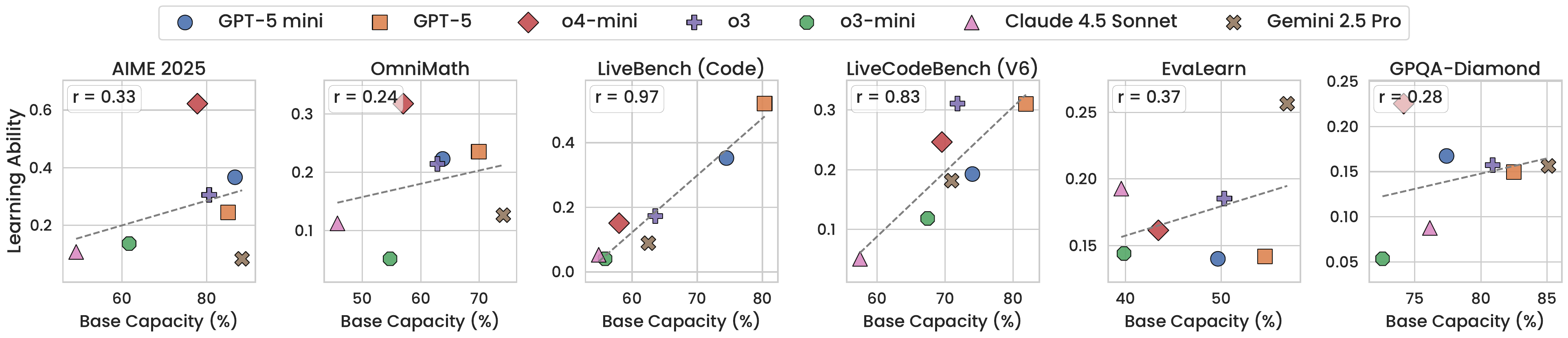}
\caption{Zero-shot performance of models (Base Capacity) and the learning gain show positive Pearson correlations (r), indicating that better-performing LLMs possess greater improving capability. 
Scores are averaged across 3 runs.}
\label{fig:corr}
\end{figure*}

\hspace{1em}
\begin{mdframed}[backgroundcolor=gray!15] 
\noindent\textbf{Findings 3}: Improving Capability: LLMs that perform better on the task shows higher learning gain during test-time.
\end{mdframed}

We calculate the improving capability of a model $\mathcal{M}$ using $\Delta_{\mathcal{M}} = \frac{S_{\text{max}} - S_{\text{base}}}{1 - S_{\text{base}}}$, where $S_{\text{base}}$ denotes the model's initial zero-shot performance without feedback and $S_{\text{max}}$ represents its peak accuracy achieved under our CoE setting with model self feedback. All numbers are averaged across three runs.
In Figure~\ref{fig:corr}, we observe a clear positive trend between base performance and improving capability across benchmarks. 
For instance, models on both coding tasks \eg, LiveBench (Code) with $r = 0.97$) and LiveCodeBench (V6) ($r = 0.83$) show strong correlations, indicating that models with stronger initial reasoning ability tend to learn more effectively from feedback. 
Although tasks like AIME 2025 ($r = 0.33$) and  OmniMath ($r = 0.24$) show relatively weaker correlations, the overall trend remains consistent, with an average task-level Pearson correlation of 0.50.
These findings suggest that learning from experience is an emergent property that scales with model capacity --- stronger LLMs are inherently better at digesting feedback and improving through interactions.

\begin{figure*}[t]
  \centering
  \includegraphics[width=.9\textwidth]{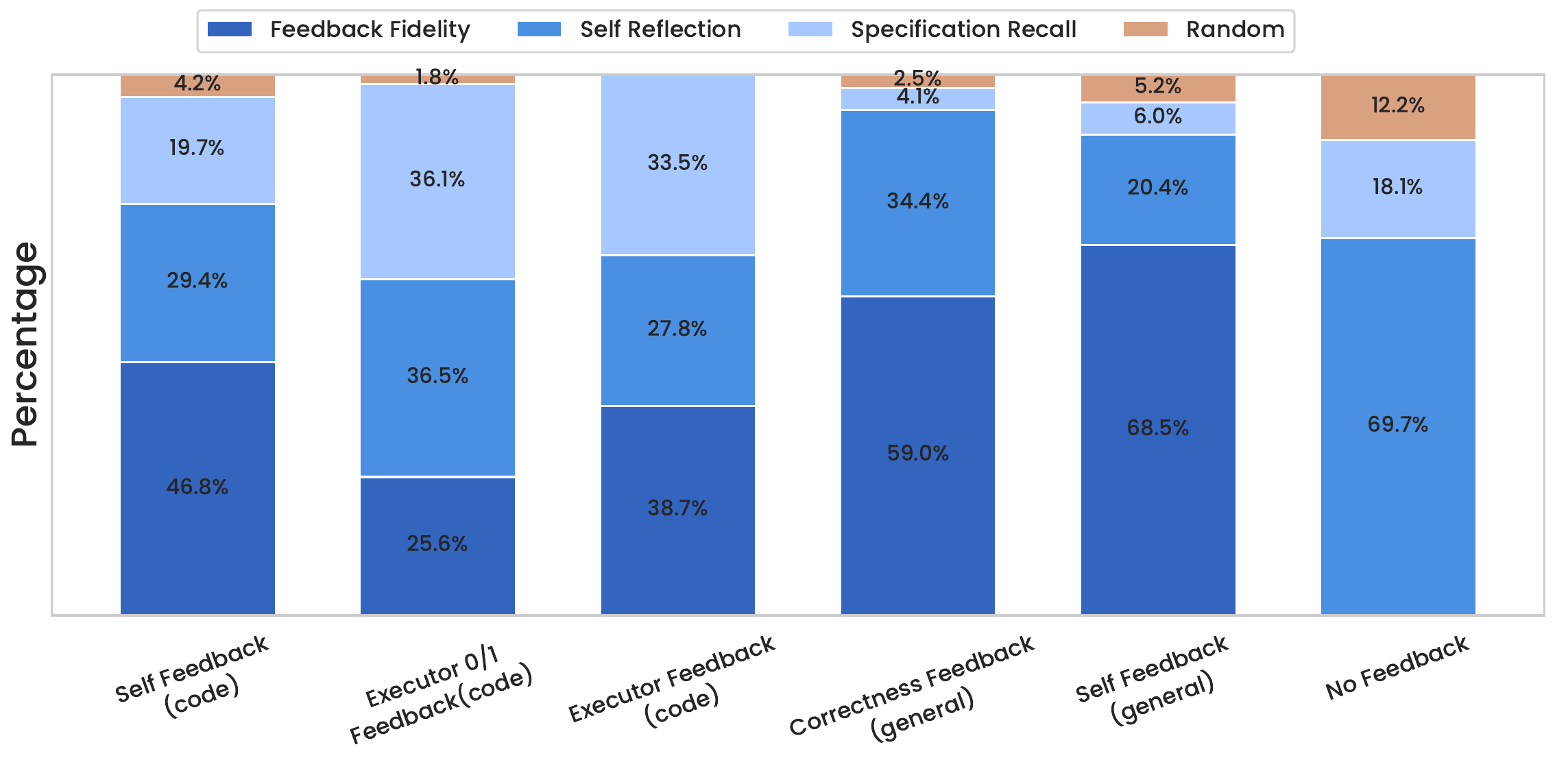}
\vspace{-1em}
\caption{Percentages of different reasons for LLMs' improvement patterns from 6,630 incorrect to correct response pairs. We employ GPT-5 to conduct this automatic analysis.}
\vspace{-1em}
\label{fig:pattern_analysis}
\end{figure*}

\section{Further Discussion and Conclusion}
\vspace{-.5em}
To further investigate LLMs’ learning capacity under CoE, we conduct ablations on the iterative problem-solving setup, including spurious feedback, improvement pattern analysis, dual feedback combinations, and principled experience selection. We additionally study model behavior under varying feedback strengths (Appendix~\ref{app:feedback_strength}) and extended experience iterations (Appendix~\ref{app:extended_iterations}).

\noindent \textbf{Learning from Spurious Feedback.}
To investigate the robustness of LLMs under spurious feedback, we design an experiment where models receive exclusively \textit{incorrect} or \textit{correct} feedback (\eg, always stating ``the answer is incorrect'' or vice versa). We report the best model performance over 20 iterations in Table~\ref{tab:spurious_label}. This setup evaluates whether models can recover from (or even benefit under) misleading feedback signals.
We find that spurious feedback generally degrades performance by average 7.6\% on AIME 2025 and 2.6\% on GPQA-Diamond. Yet, stronger models such as GPT-5 mini exhibit greater robustness, with only minor drops of 2.5\% and 0.6\%, compared to o4-mini’s declines of 12.8\% and 4.6\%.

To further enhance reliability, we introduce \textit{Selective Majority Voting} (SelMV-$n$), which aggregates final answers via majority voting among the first $n$ valid attempts.
Interestingly, on GPQA-Diamond, SelMV with incorrect feedback surpasses model feedback by 0.9\% (79.4\% $\to$ 80.3\%), while accuracies after SelMV improve by average 1.2\% and 2.3\% on AIME 2025 and GPQA-Diamond, respectively, underscoring the robustness of reasoning LLMs despite adversarial signals. Further results on spurious and different levels of feedback are in Appendix~\ref{app:spurious} and~\ref{app:feedback_strength}.

\noindent \textbf{Analysis of Improvement Patterns.}
To better understand reasons for the improvement of LLMs through iterative feedback, we design the experiments to analyze the ``why'' behind each flip from incorrect to correct answer of LLMs. We collect 6,630 examples across all tasks and five models (\ie, GPT-5, GPT-5 mini, o4-mini, o3, o3-mini) and leverage the latest GPT-5 model to analyze the cause. To validate this automated analysis, we compute Cohen's Kappa between GPT-5 and human ratings on 100 randomly sampled trajectories (25 per category), achieving 76.8\% agreement, indicating substantial agreement (details in Appendix~\ref{app:judge_agreement}). 
We define four factors behind the improvement of LLMs: Feedback Fidelity for improving from trail feedback, Self Reflection for referring to self-reflection, Specification Recall for correcting based on the question and/or format requirements, Random for model improving from other reasons. We present the detailed criteria of these factors and prompts in Appendix~\ref{app:analysis_improvement} and~\ref{app:analyze_prompt}.

In Figure~\ref{fig:pattern_analysis}, we observe: (1) 47.7\% of all improvements are feedback-driven, confirming that LLMs meaningfully interpret and act on feedback signals. (2) In coding tasks, 30.0\% of improvements stem from specification recall, reflecting the format- and syntax-sensitive nature of these tasks. (3) Model-generated feedback elicits a higher feedback-related improvement proportion than other sources (\eg, 58.7\% vs. 41.1\%), suggesting self-generated feedback is more contextually aligned.

\noindent \textbf{Dual Feedback CoE.}
To test whether multiple feedback channels are complementary, we combine model feedback with correctness (math) or executor feedback (coding) using Claude 4.5 Sonnet. As shown in Table~\ref{tab:dual_feedback}, dual feedback shows clear complementarity: on AIME 2025, dual feedback reaches 76.7\%, surpassing correctness-only (70.0\%) and model-only (60.0\%); on LiveBench (Code), dual feedback achieves 81.2\% vs. 78.1\% (executor) and 57.8\% (model). On the more challenging OmniMath, gains are marginal (73.5\% vs. 74.5\% correctness-only), with correctness signals typically pushing best rounds later. These results suggest that feedback channels contribute complementary signals, though the marginal benefit depends on task difficulty. Full results are provided in Appendix~\ref{app:dual_feedback}.

\noindent \textbf{CoE with Principled Experience Selection.}
We further investigate whether memory-based selection and compression mechanisms add benefit beyond full experience trails by combining CoE with Dynamic CheatSheet (DC)~\citep{suzgun2025dynamic} and SimpleMem~\citep{liu2026simplemem} under the same within-task protocol (no cross-task leakage). As shown in Table~\ref{tab:dual_feedback}, these methods do not outperform pure self-feedback: on AIME 2025, model feedback alone achieves 60.0\% vs. 50.0\% (+DC) and 56.7\% (+SimpleMem); on LiveBench (Code), 57.8\% vs. 51.6\% (+DC) and 54.7\% (+SimpleMem), suggesting that aggressive compression may discard critical intermediate reasoning.

\begin{table*}[t]
\centering
\small
\begin{tabular}{lcc|cc|cc}
\toprule
& \multicolumn{2}{c|}{\textbf{AIME 2025}} & \multicolumn{2}{c|}{\textbf{LiveBench (Code)}} & \multicolumn{2}{c}{\textbf{OmniMath}} \\
\textbf{Setting} & Acc & Best R & Acc & Best R & Acc & Best R \\
\midrule
Dual (Model + Corr/Exec) & \textbf{76.7} & R19 & \textbf{81.2} & R15 & 73.5 & R17 \\
Correctness / Executor & 70.0 & R13 & 78.1 & R15 & \textbf{74.5} & R17 \\
Binary Executor & -- & -- & 71.9 & R13 & -- & -- \\
Model & 60.0 & R6 & 57.8 & R17 & 50.5 & R9 \\
Model + DC & 50.0 & R8 & 51.6 & R15 & 46.0 & R10 \\
Model + SimpleMem & 56.7 & R6 & 54.7 & R17 & 49.5 & R12 \\
\bottomrule
\end{tabular}
\caption{Dual feedback and principled experience selection results using Claude 4.5 Sonnet. Dual feedback combines model feedback with correctness (math) or executor (code) signals. Memory-based methods (DC, SimpleMem) are applied within-task with no cross-task leakage. Acc: best accuracy (\%) over 20 iterations; Best R: iteration achieving best performance.}
\label{tab:dual_feedback}
\end{table*}

\begin{table}
\centering
\small
\begin{tabular}{ccccc}
\toprule
Feedback   & \multicolumn{2}{c}{AIME 2025} & \multicolumn{2}{c}{GPQA Diamond}  \\
\hline
                & \makecell{GPT-5\\mini} & \makecell{o4\\mini}        & \makecell{GPT-5\\mini} & \makecell{o4\\mini}              \\
                \hline
\textcolor{gray}{Self}  & \textcolor{gray}{93.3}       & \textcolor{gray}{91.1}           & \textcolor{gray}{79.9}       & \textcolor{gray}{78.8}                 \\
\textcolor{gray}{SelMV Self}  & \textcolor{gray}{91.1}       & \textcolor{gray}{88.9}           & \textcolor{gray}{80.4}       & \textcolor{gray}{79.5}                 \\
\hline
All \textit{Correct}     & 90.0       & 73.3           & 79.3       & 75.8                 \\
SelMV \textit{Correct}   & \textbf{93.3}       & 73.3           & 79.3       & 76.3                 \\
\textit{Incorrect}       & 91.7       & 83.3           & 79.3       & 72.7                 \\
SelMV \textit{Incorrect} & 89.7       & \textbf{86.7}           & \textbf{82.8}       & \textbf{77.8}                \\ \bottomrule
\end{tabular}

\caption{The best performance over 20 iterations under constant \textit{correct}'' or \textit{incorrect}'' feedback (\eg, ``the answer is correct''). Selective majority voting (SelMV) helps LLMs maintain performance. Results are averaged over 3 runs with best scores \textbf{emphasized}.}
\label{tab:spurious_label}
\end{table}

\noindent \textbf{Conclusion.}
We present a comprehensive analysis of Chain-of-Experience (CoE), showing that LLMs can improve during inference through iterative feedback and accumulated experience. 
Across math, coding, and knowledge tasks, methods in CoE consistently enhances performance and efficiency, demonstrating the effectiveness of feedback-driven test-time learning.
Our analysis also reveals a positive correlation between model ability and improvement capacity, and shows that most gains emerge early, even under weak or spurious feedback. Finally, we analyze various model improvement patterns during this iterative process.
\newpage

\section{Acknowledgement}
The authors sincerely thank Deyao Zhu, Shu Zhong, and Guang Shi for providing valuable feedback and discussions on the experimental part and presentation of the paper.

\bibliographystyle{plainnat}
\bibliography{colm2026_conference}
\newpage

\appendix
\section{Detailed Experimental Setup} \label{app:exp_setup}

\textbf{Different Reasoning Efforts.}
For OpenAI models, we employ the default `low' and `high' in the $\text{reasoning}\_\text{effort}$ parameter to twitch models' reasoning level. For Claude 4.5 Sonnet, we disable the thinking mode and set the thinking budget to 10,000 as the low and high reasoning variants, respectively.

\textbf{Decoding Parameters.}
For OpenAI models, we use the default decoding parameter with temperature set to 1.0. For Gemini, Claude series, we use a temperature of 0.2 for decoding.

\section{Averaged Results}
\label{app:avg_main_results}
We present the averaged best scores over 20 iterations of different methods in Table~\ref{tab:feedback_comparison_full}. From the table, we can clearly draw the conclusion that feedback helps LLMs perform better on all six benchmarks, while existing self-improving algorithms (\ie, ACE, DC) do not perform decently on this testing suite.
\begin{table*}[t]
\centering
\scriptsize
\begin{tabular}{lcccccc}
\toprule
\textbf{Method} 
& \textbf{AIME 2025} 
& \textbf{\makecell[c]{LiveCodeBench\\(V6)}} 
& \textbf{\makecell[c]{LiveBench\\(Code)}} 
& \textbf{OmniMath} 
& \textbf{\makecell[c]{GPQA\\Diamond}} 
& \textbf{EvaLearn} \\
\midrule
ICL                & 71.83\% & 62.50\% & 65.46\% & 53.12\% & 78.45\% & 40.99\% \\
ACE                & 71.98\% & 66.94\% & 69.38\% & 50.33\% & 76.58\% & 42.54\% \\
DC                  & 73.33\% & 63.59\% & 68.58\% & 48.64\% & 79.56\% & 42.68\% \\
w/o Feedback       & 77.78\% & 72.57\% & 60.16\% & 65.17\% & 80.02\% & 44.91\% \\
Reasoning-high             & 69.05\% & 70.63\% & 55.46\% & 61.81\% & 76.21\% & 39.58\% \\
Reasoning-low             & 60.48\% & 61.03\% & 55.38\% & 50.60\% & 72.92\% & 29.34\% \\
Binary-Executor         & --      & 72.90\% & 71.65\% & --      & --      & --      \\
Self       & 82.22\% & \textbf{75.69\%} & 69.94\% & 67.52\% & 81.03\% & 51.73\% \\
Correctness/Executor      & \textbf{89.05\%} & 74.50\% & \textbf{75.78\%} & \textbf{79.61\%} & \textbf{99.52\%} & \textbf{57.05\%} \\
\bottomrule
\end{tabular}
\caption{Average performance comparison (\%) across different LLMs on different datasets. For baselines, ICL, ACE, DC stands for few-shot in-context learning, agentic context engineering, and dynamic cheatsheet, respectively. }
\label{tab:feedback_comparison_full}
\end{table*}

\section{Evaluated Benchmarks} \label{app:eval_bmks}
We present detailed descriptions of our evaluated benchmarks below:
\begin{itemize}
    \item \textcolor{math}{AIME 2025}~\citep{balunovic_srimatharena_2025} is a challenging and universal math benchmark consists of 30 cases from AIME in 2025. 
    \item \textcolor{math}{OmniMath}~\citep{gao2024omni} is a universal olympiad level mathematic benchmark consists of a total of 4,428 problems, and we sample 200 examples across all difficulties for evaluation.
    \item \textcolor{coding}{LiveCodebench (V6)}~\citep{jain2024livecodebench} is an evolving benchmark for challenging code generation. We select its latest version (V6) alone with a total of 175 samples.
    \item \textcolor{coding}{LiveBench (Code)}~\citep{white2024livebench} is originated from a living benchmark spans across six aspects. We select the coding aspect, comprising 128 samples for evaluation.
    \item \textcolor{knowledge}{EvaLearn}~\citep{dou2025evalearn} is the first benchmark that evaluate the experience learning abilities of language models, which includes total 648 examples.
    \item \textcolor{knowledge}{GPQA Diamond}~\citep{rein2024gpqa} is a challenging multiple-choice question set in biology, chemistry, and physics, authored by PhD-level experts. It consists of 198 examples in total.
\end{itemize}

\section{Baselines} \label{app:baselines}

\textbf{Built-in Reasoning}: A native test-time scaling mechanism implemented in OpenAI models allows the reasoning effort to be adjusted between low and high. We treat this as a built-in and straightforward scaling baseline and present further details in Appendix~\ref{app:exp_setup}

\textbf{Few-shot In-Context Learning (ICL).}
In few-shot ICL, each demonstration is a previously solved question–answer pair. As the model processes tasks sequentially, all past pairs are stored in an experience buffer. For a new question, we retrieve the $k$ most similar past questions using embeddings from OpenAI’s \texttt{text-embedding-3-large} model~\citep{openai2024_embedding}, and include their original question–answer formats as demonstrations, followed by the new question. When fewer than $k$ examples are available, all prior examples are used. We evaluate ICL with $k\in [1, 5, 8, 12, 15, 20]$.

\textbf{Dynamic CheatSheet (DC)~\citep{suzgun2025dynamic}.}
Dynamic CheatSheet (DC) is a test-time learning method that maintains an adaptive external memory of reusable strategies or code snippets distilled from prior solutions. As new problems are solved, DC summarizes high-level strategies from selected past tasks—using ground-truth answers for clean experience curation—and stores them as structured cheatsheets. Past tasks are retrieved using the same similarity-based retrieval as few-shot ICL; when fewer than $k$ tasks exist, all available examples are used. Consistent with ICL, cheatsheets are synthesized from the most recent $k\in [1, 5, 8, 12, 15, 20]$ relevant solutions.

\textbf{Agentic Context Engineering (ACE)~\citep{zhang2025agentic}.}
Agentic Context Engineering (ACE) is a context adaptation framework that treats prompts as evolving playbooks rather than static demonstrations. It incrementally distills reusable strategies and domain insights through a generate–reflect–curate process, producing localized context updates that preserve prior knowledge and avoid monolithic rewrites. Similar to few-shot ICL, ACE retrieves relevant past trajectories and integrates insights from the most recent $k\in [1, 5, 8, 12, 15, 20]$ trajectories into its evolving playbook.

\section{Benchmark Input and Answer Examples} \label{app:benchmark_examples}
We provide representative input and answer examples for three benchmarks below.

\noindent \textbf{AIME 2025.}
\begin{tcolorbox}[colback=gray!5!white, colframe=gray!75!black, boxsep=1mm, left=1mm, right=1mm, top=1mm, bottom=1mm]
{\small
\texttt{Solve the following AIME problem step by step and provide the final answer:} \\
\texttt{$2^{x} + 2^{-x} = 5$.} \\
\texttt{Compute $2^{2x} + 2^{-2x}$.} \\
\texttt{Answer format: ANSWER: <integer>} \\[0.5em]
\texttt{Answer: 21}
}
\end{tcolorbox}

\noindent \textbf{GPQA Diamond.}
\begin{tcolorbox}[colback=gray!5!white, colframe=gray!75!black, boxsep=1mm, left=1mm, right=1mm, top=1mm, bottom=1mm]
{\small
\texttt{Which of the following is *not* a property of an ideal gas under standard conditions?} \\
\texttt{A) Molecules occupy negligible volume} \\
\texttt{B) Collisions are perfectly elastic} \\
\texttt{C) Internal energy depends only on temperature} \\
\texttt{D) Pressure is independent of volume} \\[0.5em]
\texttt{Answer: D}
}
\end{tcolorbox}

\noindent \textbf{LiveCodeBench (V6).}
\begin{tcolorbox}[colback=gray!5!white, colframe=gray!75!black, boxsep=1mm, left=1mm, right=1mm, top=1mm, bottom=1mm]
{\small
\texttt{Question: "Erase Leaves --- Given a tree with N vertices, repeatedly remove leaf vertices. Find the minimum operations to delete vertex 1."} \\[0.3em]
\texttt{Public: [\{"input": "9\textbackslash n1 2\textbackslash n2 3\textbackslash n...", "output": "5\textbackslash n"\}, ...]} \\
\texttt{Private: [\{"input": "2\textbackslash n1 2\textbackslash n", "output": "1\textbackslash n"\}, ...]}
}
\end{tcolorbox}

\noindent We use exact match for AIME 2025 and GPQA-Diamond, LLM-as-a-Judge for OmniMath~\citep{gao2024omni}, and a Python interpreter for coding tasks (LiveBench-Code and LiveCodeBench).

\section{Full Discussions} \label{app:full_discussion}
In this section, we present the full version of different discussions in 
\subsection{LLM with All Spurious ``Correct'' Feedback} \label{app:spurious}
\begin{figure*}[h]
  \centering
  \includegraphics[width=.65\textwidth]{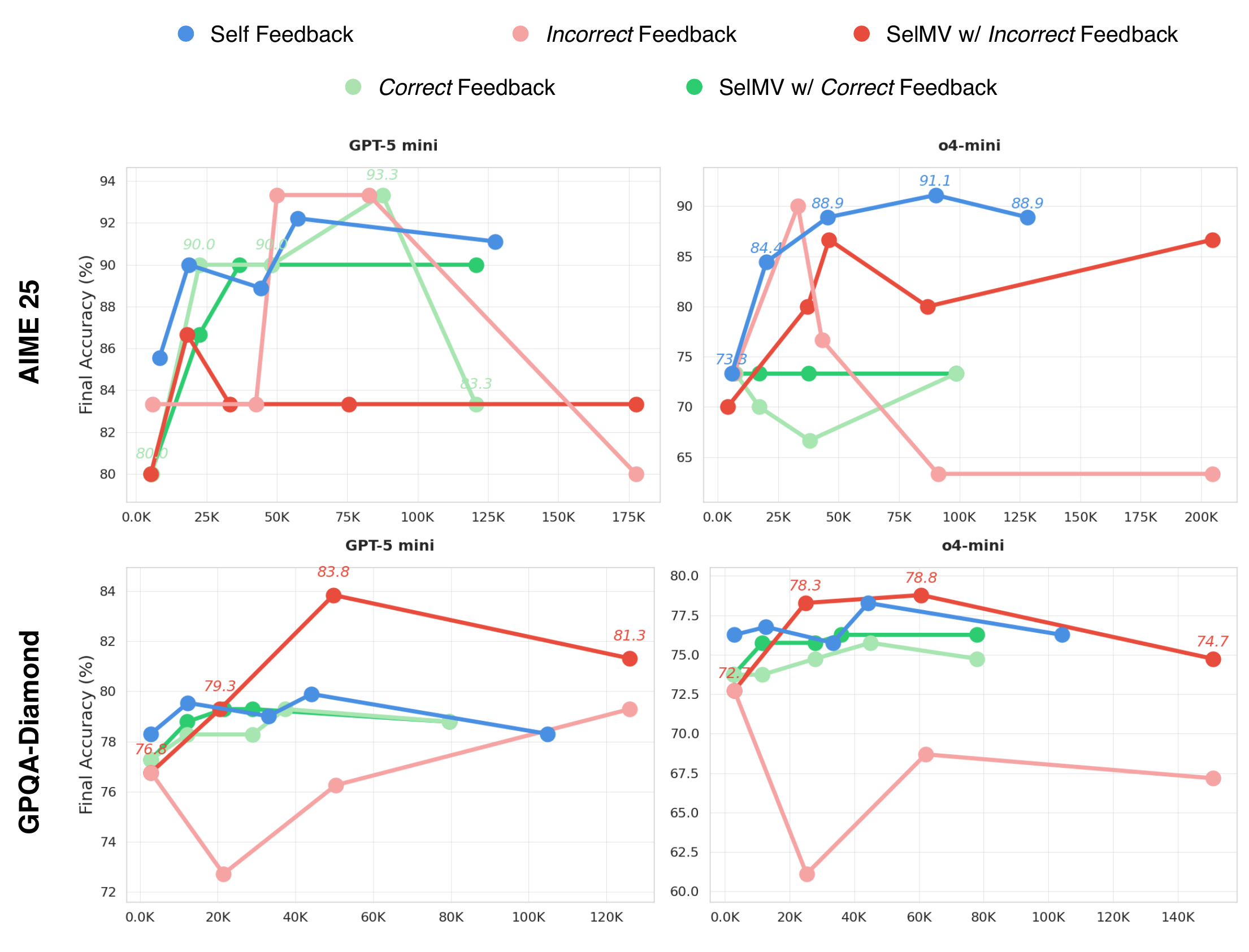}
\caption{Model performance using constant ``incorrect'' and ``correct'' feedback. By leveraging the selective majority voting, LLMs show decent performance when facing spurious feedback on math tasks.}
\label{fig:full_spurious}
\end{figure*}

In Figure~\ref{fig:full_spurious}, we further present model performance under two extreme conditions: receiving uniformly ``correct'' feedback (\eg, the answer is correct) and the SelMV-augmented results. 
We observe that although performance initially drops after exposure to such incorrect feedback, which suggests temporary confusion in adapting to inconsistent supervision. The models quickly recover and even improve as they adapt to the underlying pattern. 
Interestingly, both GPT-5 mini and o4-mini exhibit larger gains when exposed to entirely incorrect feedback, as such feedback compels the models to re-evaluate their reasoning and verify their outputs. In contrast, consistently positive feedback tends to induce overconfidence, misleading the models into accepting their initial responses without critical reassessment. This observation suggests that, paradoxically, constructive noise (in the form of seemingly negative feedback) can sometimes stimulate deeper reasoning and enhance robustness in iterative test-time learning.

\subsection{Extended Rounds of Iterations}
\label{app:extended_iterations}
\begin{figure*}[t]
  \centering
  \includegraphics[width=\textwidth]{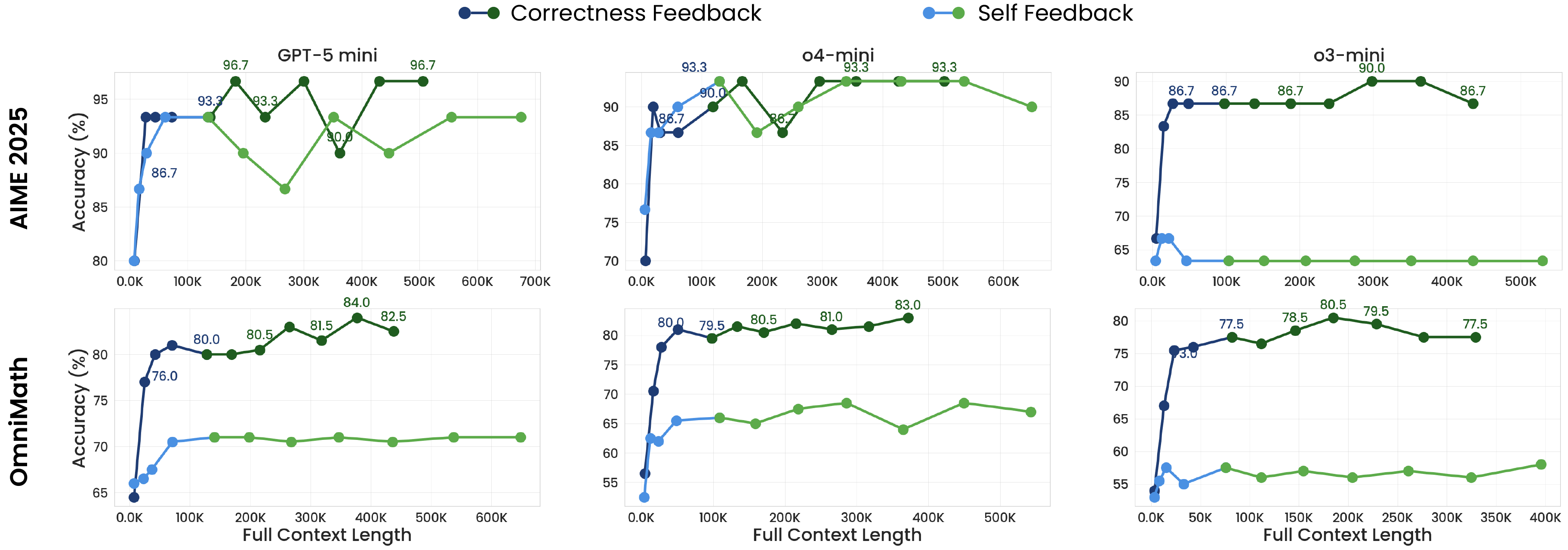}
\caption{Performance of GPT-5 mini, o4-mini, and o3-mini with extended iterations of experience to 50 on math tasks. We mark performance points within the first 20 iterations of experience in blue, and those from iterations 20 to 50 in green.}
\label{fig:extend_iter}
\end{figure*} 

To further probe the learning capacity of LLMs, we extend the number of experience iterations from 20 to 50, as shown in Figure~\ref{fig:extend_iter}. 
Across both AIME 25 and OmniMath, we observe that most performance gains occur within the first 20 iterations, while later stages yield only marginal improvements (\eg, average 16.7\% $>$ 2.2\% on AIME 25; 21.2\% $>$ 3.5\% on OmniMath). 
This trend consistently holds across different models, suggesting that LLMs quickly internalize and consolidate the useful feedback signals in the early stages, after which learning saturates. 
These results highlight that the majority of test-time learning under CoE happens rapidly --- shorter adaptation loops in our CoE are sufficient for most reasoning tasks, with less significant returns from prolonged experience accumulation.

\subsection{Feedback Strength}
\label{app:feedback_strength}
\begin{figure*}[t]
  \centering
  \includegraphics[width=\textwidth]{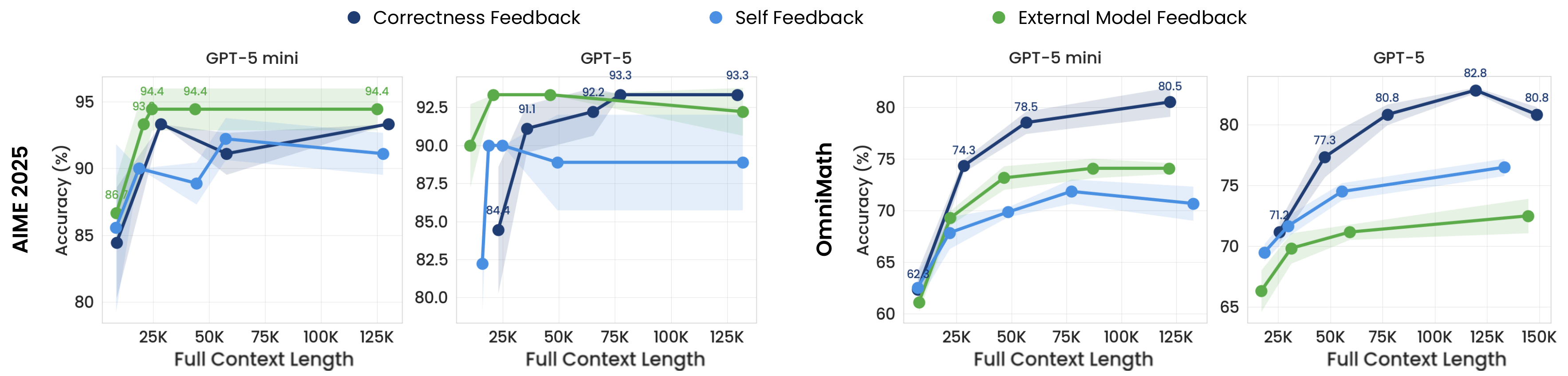}
\caption{Model performance of GPT-5 and GPT-5 mini with external model feedback on two math tasks.}
\label{fig:verifier}
\end{figure*}
For experience with model feedback, one intuitive exploration is to design CoE with different feedback providers. 
Specifically, we cluster GPT-5 and GPT-5 mini as a pair and allow each to serve as a feedback generator for the other on a mathematical and a coding task, as illustrated in Figure~\ref{fig:verifier}. 
Interestingly, external model feedback proves to be highly effective on these two benchmarks, where it even slightly surpasses correctness-based/executor feedback.
For instance, GPT-5 mini achieves a peak accuracy of 94.4\% with model feedback, compared to 93.0\% with correctness feedback; similarly, GPT-5 reaches 93.3\% vs. 92.2\%, showing that high-quality model-generated judging signals can substitute explicit correctness supervision.
On the more challenging OmniMath benchmark, GPT-5 continues to help its mini variant outperform its self-feedback baseline. 
However, because GPT-5 mini starts with only 62.5\% zero-shot accuracy, it is unable to provide sufficiently reliable feedback to improve GPT-5, resulting in inferior performance relative to correctness feedback. Moreover, both GPT-5 and GPT-5 mini underperform their AIME 2025 results ($\geq$60\% vs. $\geq$80\%), and correctness feedback remains clearly superior: GPT-5 gets 82.8\% with correctness feedback but only 74.5\% with external model feedback.
Overall, these findings highlight a consistent trend: feedback quality correlates with the verifier’s base ability on the task, and model-generated feedback becomes competitive with correctness supervision only when the verifier is sufficiently strong, suggesting a practical threshold for deploying model-as-judge in iterative experience frameworks.

\subsection{Analysis of Improvement Patterns} \label{app:analysis_improvement}
We present the detailed criteria that we used for classifying the 

\begin{itemize}
    \item Feedback Fidelity: Improvements directly driven by external feedback, where the model explicitly incorporates provided guidance or corrections into its next response.
    \item Self Reflection: Improvements arising from the model’s own reasoning, identifying and correcting errors with little reliance on external feedback.
    \item Specification Recall: Adjustments motivated by task instructions or formatting requirements, as the model re-aligns with the original question or output schema.
    \item Random: Changes with no identifiable cause, typically minor rewording or stylistic variations unrelated to feedback or specification.
\end{itemize}

\subsection{Human--GPT Judge Agreement} \label{app:judge_agreement}
To validate the GPT-5-based automatic improvement analysis, we randomly sample 100 incorrect-to-correct trajectory pairs (25 per category: Feedback Fidelity, Self Reflection, Specification Recall, Random) from the full set of 6,630 examples. Two human annotators independently classify each pair into the four categories using the same criteria as the GPT-5 judge (Appendix~\ref{app:analyze_prompt}). We then compute Cohen's Kappa between the GPT-5 labels and the majority human label.

\begin{table}[h]
\centering
\small
\begin{tabular}{lcc}
\toprule
\textbf{Category} & \textbf{Agreement (\%)} & \textbf{Cohen's $\kappa$} \\
\midrule
Feedback Fidelity & 84.0 & 0.81 \\
Self Reflection & 72.0 & 0.71 \\
Specification Recall & 80.0 & 0.78 \\
Random & 68.0 & 0.63 \\
\midrule
\textbf{Overall} & \textbf{76.0} & \textbf{0.768} \\
\bottomrule
\end{tabular}
\caption{Cohen's Kappa agreement between GPT-5 judge and human annotators across four improvement categories on 100 sampled trajectories. The overall $\kappa$ of 0.768 indicates substantial agreement~\citep{landis1977measurement}.}
\label{tab:judge_agreement}
\end{table}

\noindent The overall $\kappa$ of 0.768 falls in the ``substantial agreement'' range, confirming that GPT-5 is a reliable proxy for human attribution in this task. Disagreements concentrate in the Random and Self Reflection categories, where the distinction between stochastic drift and genuine self-correction can be ambiguous even for human raters.

\section{BrowseComp-Plus} \label{app:browsecomp}
We also report model performance on BrowseComp-Plus~\cite{chen2025browsecomp}. 
It is a benchmark to evaluate deep research systems, isolating the effect of the retriever with a local database. It is sourced from the BrowseComp~\cite{wei2025browsecomp} and we sample 200 examples to accelerate the evaluation.

\begin{figure*}[h]
  \centering
  \includegraphics[width=\textwidth]{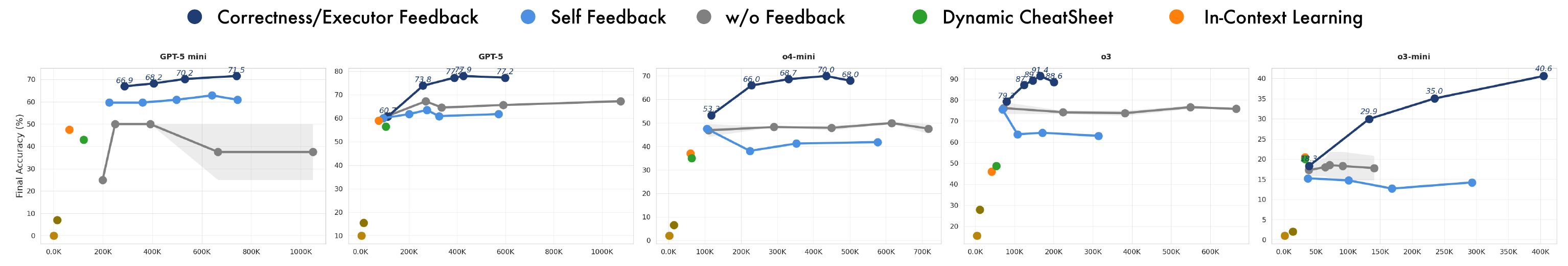}
\caption{On BrowseComp-Plus, self-feedback models fall behind as the task requires external search-based knowledge.}
\label{fig:browsecomp}
\end{figure*}

In Figure~\ref{fig:browsecomp}, we observe that unlike coding and math tasks, BrowseComp-Plus requires knowledge beyond the scope of the models’ training data. Consequently, for most models, incorporating \textcolor{selffeedcolor}{self feedback} leads to a performance decline compared to the no-feedback setting, highlighting the limitation of relying solely on self feedback in out-of-distribution knowledge scenarios.

\section{Full Results of Performance and Efficiency} \label{app:full_price}
\begin{figure*}[h]
  \centering
  \includegraphics[width=\textwidth]{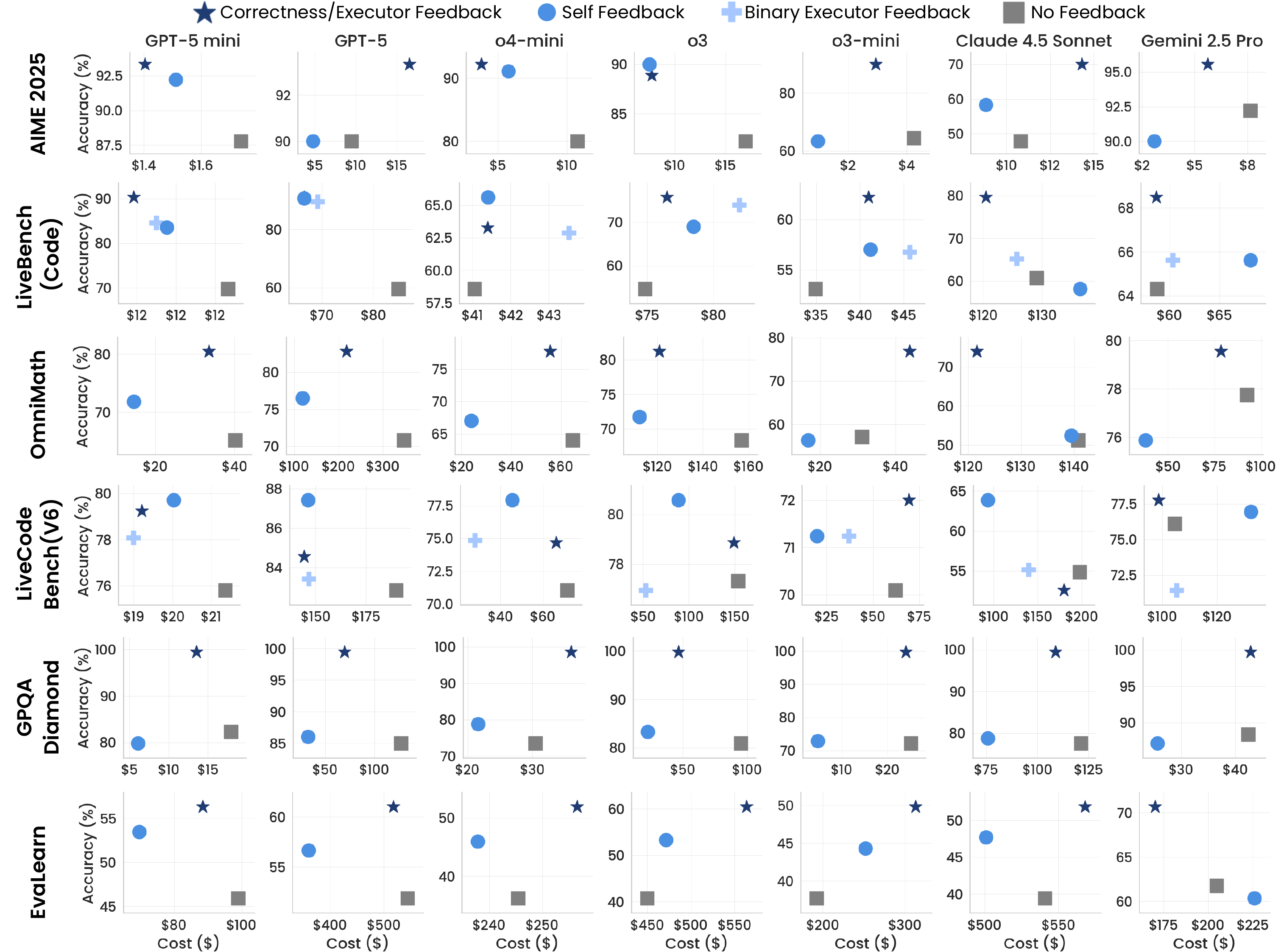}
\caption{Full results of total API cost (in dollar) vs. best model performance over 20 iterations. LLMs with detailed feedback (\ie, \textcolor{selffeedcolor}{self feedback}) achieves decent results with fewer costs, while CoE without feedback generally falls behind (at lower right). }
\label{fig:full_price}
\end{figure*}

We present full results of model performance (Figure~\ref{fig:main}) and API costs (Figure~\ref{fig:full_price}) regarding eight reasoning LLMs over six benchmarks. 
\begin{figure*}[t]
  \centering
  \includegraphics[width=\textwidth]{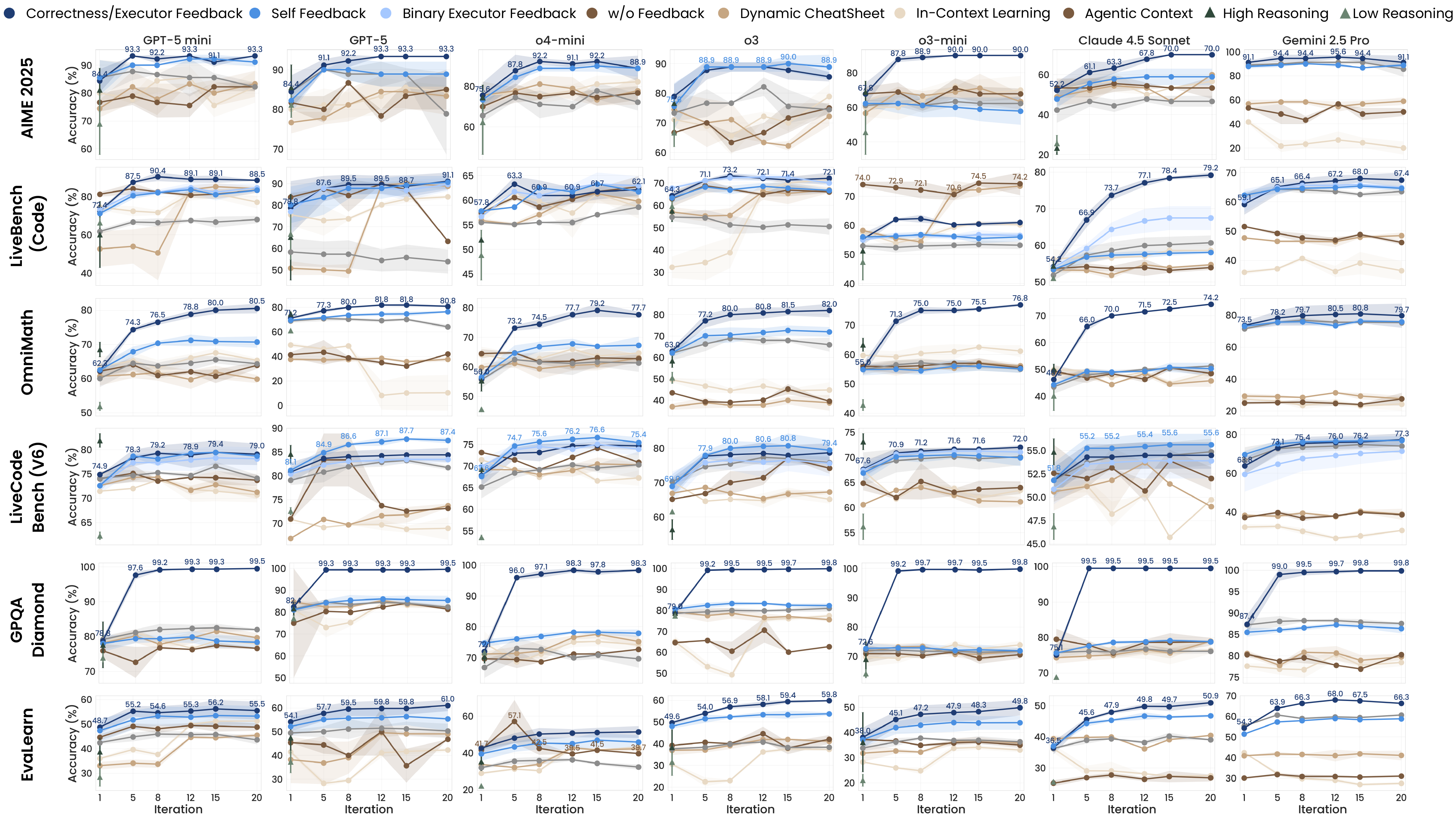}
\caption{Full results of eight state-of-the-art LLMs on six benchmarks incorporating different generation techniques. Model accuracies with different levels of feedback (\textcolor{feedbackcolor}{correctness/executor}, \textcolor{selffeedcolor}{self}, and \textcolor{bexefeedbackcolor}{binary executor} feedback) generally perform better than the baseline strategies.}
\label{fig:main_full}
\end{figure*}

\section{Token Analysis} \label{app:token_analysis}
We report aggregated token counts alongside accuracy for representative methods in Table~\ref{tab:token_performance}. CoE with feedback achieves higher accuracy at comparable token budgets to other multi-round methods, demonstrating that feedback-driven iterations yield higher return per token rather than simply inflating prompt length.

\begin{table}[h]
\centering
\small
\begin{tabular}{llrc}
\toprule
\textbf{Dataset} & \textbf{Method} & \textbf{Tokens} & \textbf{Acc (\%)} \\
\midrule
\multirow{4}{*}{AIME 2025} & CEF & 108,734 & \textbf{84.6} \\
 & SF & 108,231 & 83.8 \\
 & NF & 106,825 & 74.1 \\
 & DC & 11,233 & 74.7 \\
\midrule
\multirow{4}{*}{OmniMath} & CEF & 176,412 & \textbf{74.2} \\
 & SF & 175,806 & 72.1 \\
 & NF & 173,944 & 66.8 \\
 & DC & 16,904 & 63.9 \\
\midrule
\multirow{4}{*}{\makecell[l]{LiveCodeBench\\(V6)}} & CEF & 224,118 & \textbf{72.6} \\
 & SF & 223,441 & 71.2 \\
 & NF & 221,550 & 68.0 \\
 & DC & 20,771 & 66.4 \\
\bottomrule
\end{tabular}
\caption{Token complexity vs. accuracy across methods. CEF: Correctness/Executor Feedback; SF: Self Feedback; NF: No Feedback; DC: Dynamic CheatSheet. Token counts are aggregated across all iterations.}
\label{tab:token_performance}
\end{table}

\section{Dual Feedback and Principled Selection} \label{app:dual_feedback}
We present full results of dual feedback CoE and principled experience selection experiments using Claude 4.5 Sonnet. In the dual feedback setting, model feedback is combined with correctness feedback (math tasks) or executor feedback (coding tasks) within each iteration. For principled selection, we integrate Dynamic CheatSheet (DC)~\citep{suzgun2025dynamic} and SimpleMem~\citep{liu2026simplemem} within the same task (no cross-task retrieval) to test whether memory-based compression adds benefit beyond full experience trails.

Dual feedback demonstrates clear complementarity on AIME 2025 and LiveBench (Code), where combining two feedback channels surpasses either channel alone. On the more challenging OmniMath, correctness feedback alone matches dual feedback, suggesting that when the primary signal is already strong, the additional model feedback provides marginal benefit. Memory-based selection methods (DC, SimpleMem) consistently underperform pure model feedback, indicating that aggressive summarization or retrieval may discard critical intermediate reasoning steps that full experience trails preserve.

\newpage
\section{Prompt for Improvement Pattern Analysis} \label{app:analyze_prompt}

\begin{tcolorbox}[
    colback=gray!5!white,
    colframe=gray!75!black,
    breakable,
    enhanced,
    boxsep=1mm,
    left=1mm,
    right=1mm,
    top=1mm,
    bottom=1mm,
    listing only,
    listing options={
        language={},
        basicstyle=\ttfamily\scriptsize,
        breaklines=true,
        columns=fullflexible,
        keepspaces=true,
        showstringspaces=false
    }
]
DUAL-AXIS LIFT/CHANGE ATTRIBUTION JUDGE

You are a **dual-axis attribution judge**.
Your job is to identify both **why** the model changed and **what** specifically changed between two consecutive attempts on the same problem.
Do **not** decide if the solution is correct overall — correctness labels are provided.
Instead, attribute the observed change along two orthogonal dimensions:

* **Change Driver (WHY)** – the motivation or trigger behind the change.
* **Change Manifestation (WHAT)** – the concrete locus or type of modification made.

---

I. Change Driver (WHY the change occurred)

These categories capture *the source or motivation* of the update in the second attempt.

1. **Feedback Fidelity** – The model directly *uses* the provided feedback to modify its output.
   *Signals:* Edits match failing test or critique location; added clause mirrors feedback.

2. **Self-Reflection / Internal Reasoning** – The model self-identifies an error or improvement without (using) explicit feedback or tests a different approach to see if it performs better.
   *Signals:* “I realized…,” “previously I miscalculated…,” or internally consistent reformulation not prompted by feedback or when there's no feedback.

3. **Specification Recall / Compliance Awareness** – The model remembers or re-aligns with task instructions or formatting requirements.
   *Signals:* Adds “FINAL ANSWER:” wrapper, adheres to requested schema, restores omitted step explicitly mentioned in prompt.

4. **Random / Drift / Unknown Driver** – The motivation cannot be inferred; the change appears stochastic or stylistic.
   *Signals:* Superficial rewording, minor ordering changes, no logical link to feedback or instruction.

---

II. Change Manifestation (WHAT changed)

These categories describe *the technical or structural form* of the change between attempts.

A. **Structural Plan / Algorithm Revision** – A new high-level approach or reformulation (e.g., brute force → DP, heuristic → formula).
*Signals:* Rewritten main structure, new helper functions, change in complexity or data representation.*

B. **Local Step Soundness \& Invariant Fix** – Correction of a local logic, variable, arithmetic, or invariant violation while keeping the overall plan.
*Signals:* Fixed off-by-one in loop, corrected variable update, repaired algebraic derivation.*

C. **Edge / Boundary Condition Handling** – Added or fixed guard for extreme/special cases (e.g., empty, zero, overflow, tie).
*Signals:* `if n == 0`, `<=` `<`, added epsilon, handled `len == 1`.*

D. **Output / Format Compliance** – Adjusted presentation or output schema without changing algorithmic content.
*Signals:* Added “FINAL ANSWER:,” fixed JSON/CSV layout, printing only required token.*

E. **Other / Ambiguous Change** – Cannot clearly assign to A–E or insufficient evidence.
*Signals:* Stylistic edits, reordering, or unrelated cleanup.*

---
... ...

VII. Constraints

* Do **not** recompute correctness or logic.
* Focus on identifying the **causal link (WHY)** and the **technical locus (WHAT)** of the change.
* Be concise: cite only minimal evidence sufficient to justify your judgment.
* If code is long, highlight the 1–2 lines most diagnostic of change.
* If feedback is visible but not followed, set `"visible": true, "operationalized": false`.
* Always describe the *learning pattern* observed in Attempt\_t+1.
\end{tcolorbox}

\section{Limitations}
Our evaluation focuses primarily on math, knowledge, and coding benchmarks. While these domains offer controlled settings to probe iterative improvement, there are interaction-intensive scenarios where experience unfolds over long horizons~\cite{jimenez2023swe,yao2024tau}, to which the CoE paradigm should naturally extend.
In addition, we do not update model parameters in this study. This choice isolates the Chain-of-Experience mechanism as a test-time paradigm, but it also means that the observed improvements arise from contextual reuse of experience rather than true learning; incorporating weight updates to internalize experience remains an important next step toward training models with persistent the ``learning-from-experience'' abilities.

\section{Declaration of AI Tool Usage}
During the preparation of this manuscript, we used OpenAI's GPT-5 model for minor language refinement and smoothing of the writing. 
The AI tool was not used for generating original content, conducting data analysis, or formulating core scientific ideas. 
All conceptual development, experimentation, and interpretation were conducted independently without reliance on AI tools.

\end{document}